# ECOLANG

## Communications Language for Ecological Simulations Network

*Messages and Communications Protocol for EcoSimNet Framework*

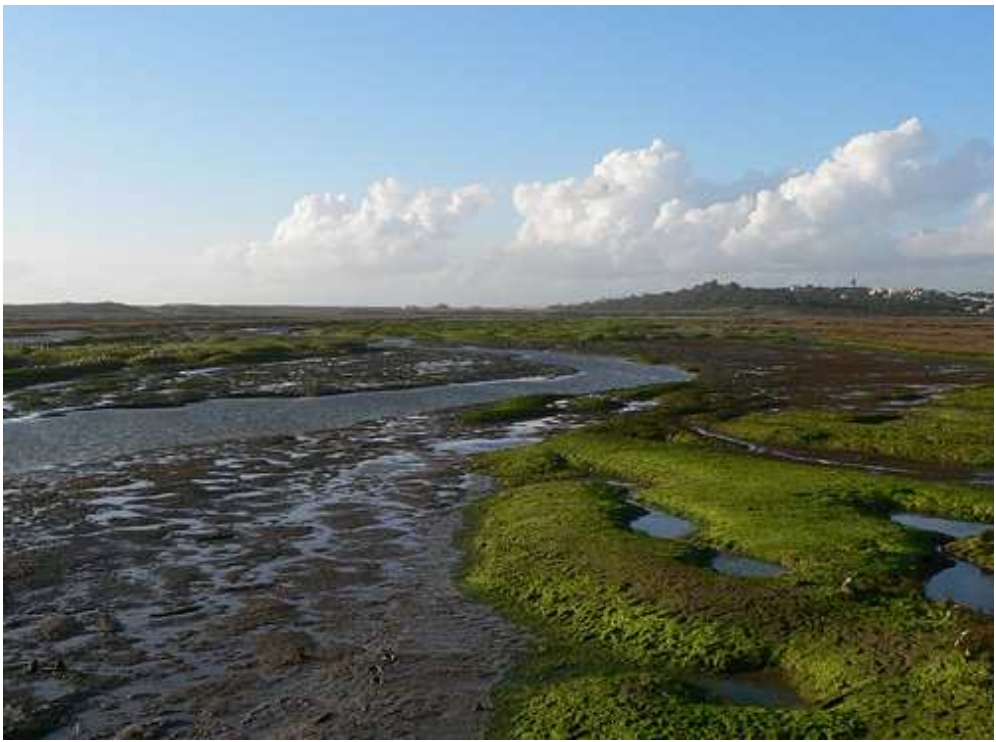

*António Pereira*

**Faculty of Engineering of University of Porto**
*LIACC – Artificial Intelligence and Computer Science Laboratory*

# Index

# Introduction

This document ("ECOLANG_v_1_3c_Eng.doc") describes the communication language used in one multi-agent systems environment for ecological simulations, based on the EcoDynamo simulator application (Pereira and Duarte 2005) linked with several intelligent agents and visualisation applications and extends the initial definition of the language (Pereira et al. 2005).

The agents' actions and perceptions are translated into messages exchanged with the simulator application and other agents.

The concepts' definitions used follow the BNF notation (Backus et al. 1960) and it's inspired in the Coach Unilang language (Reis and Lau 2002).

ECOLANG notation is an extension to the original BNF formalism adding the following meta-symbols:

> { } used for repetitive items (one or more times);
>
> [ ] encloses types of values;
>
> Terminal symbols use bold face letters.

# 1 Message definition

The base syntax of each message is as follows:

```
<MESSAGE> ::= message (<ID> <SENDER> <RECEIVER> <MSG_CONTENT>)
<ID> ::= [integer]
<SENDER> ::= [string]
<RECEIVER> ::= [string]
```

`<ID>` is the message identifier sent by the initiator – it is a sequential integer number controlled by each sender (initial value is 1).

`<SENDER>` is the name of the message initiator application (source).

`<RECEIVER>` is the name of the message destination application.

`<MSG_CONTENT>` is the content of the message.

Each message will be represented by a numeric reference to facilitate its identification.

# 2 Message types

Message exchanged by the applications may belong to four types: connection, definitions, actions and perceptions.

The connection messages establish the communication sessions between the agents / applications and specify the computer where agent belongs and the port number in which the agent listens the messages.

Though there could be actions and perceptions dedicated to some type of the agents belonging to the system, there are no restrictions to the messages that applications can use.

```
<MSG_CONTENT> ::= <CONNECTION_MSG> | <DEFINITION_MSG> | <ACTION_MSG> |
        <PERCEPTION_MSG>
```

## 2.1 Connection messages

Connection messages define the start and the finish of the communications sessions between applications. In this group there are also messages to ask the agents known by the other partner of the session.   This allows the establishment of links between multiple applications, facilitating the expansion of the communications and knowledge network.

```
<CONNECTION_MSG> ::= <CONNECT> | <DISCONNECT> | <ACCEPT> | <ASK_AGENTS> |
      <KNOWN_AGENTS>
```

The session begins with the *connect* message and finishes with the *disconnect* message. The *accept* message answers any of those messages:

```
<CONNECT> ::= connect <HOST_NAME> <HOST_ADDR> <SERVER_PORT>                [1.1]

<DISCONNECT> ::= disconnect                                                [1.2]

<ACCEPT> ::= accept (<ACTION_ID> <ACTION_RESULT>)                          [1.3]

<HOST_NAME> ::= [string]

<HOST_ADDR> ::= [string]

<SERVER_PORT> ::= [integer]

<ACTION_ID> ::= <ID>

<ACTION_RESULT> ::= ok | failed
```

To ask the other partner for the known agents:

```
<ASK_AGENTS> ::= agents                                                    [1.4]

<KNOWN_AGENTS> ::= known_agents (<ACTION_ID> {<AGENT>})                     [1.5]

<AGENT> ::= (<AGENT_NAME> <HOST_NAME> <HOST_ADDR> <SERVER_PORT> <CONNECTED>)

<AGENT_NAME> ::=[string]

<CONNECTED> ::= connected | disconnected
```

`<HOST_NAME>` and `<HOST_ADDR>` identify the computer name and IP address where the application runs, `<SERVER_PORT>` identifies the port number where the application expects connections.

`<ACTION_ID>` of the answer messages must be the `<ID>` of the corresponding connection message.

## 2.2 Definitions

From version 1.3 of the protocol, these messages include the definition of regions and information about the type of model loaded in the simulator, its dimensions, its morphology and species of shellfish and molluscs prevailing in it.

```
<DEFINITION_MSG> ::= <REGIONS_MSG> | <MODEL_DEFINITIONS>
```

Each region is referred to by name, will be of one type and will cover a certain area. It may also be defined at the expense of other regions already defined.

```
<REGIONS_MSG> ::= <DEFINE_ACTION> | <DELETE_ACTION> | <GET_REGIONS> | <GET_REGION>
      | <DEFINE_RESULT> | <DELETE_RESULT> | <GET_REGIONS_RESULT> |
      <GET_REGION_RESULT>

<DEFINE_ACTION> ::= define (<REG_NAME> <REGION>)                          [2.1]

<DELETE_ACTION> ::= delete {<REG_NAME>}                                   [2.2]

<REG_NAME> ::= [string]

<REGION> ::= <REGION_TYPE> <REGION_AREA> | {<REG_NAME>}
```

Each region is of the type land or water and, in the later, is characterized by its quality as well as the type and quality of its sediment.

```
<REGION_TYPE> ::= <LAND_REGION> | <WATER_REGION>

<LAND_REGION> ::= land

<WATER_REGION> ::= <WATER_CARACT> <SEDIMENT_CARACT>

<WATER_CARACT> ::= <SUB_INTERTIDAL> <WATER_QUALITY>

<SUB_INTERTIDAL> ::= subtidal | intertidal

<WATER_QUALITY> ::= <QUAL_SCALE>

<QUAL_SCALE> ::= excellent | good | poor

<SEDIMENT_CARACT> ::= (<SEDIMENT_TYPE> <SEDIMENT_QUALITY>)

<SEDIMENT_TYPE> ::= sandy | sand_muddy | muddy

<SEDIMENT_QUALITY> ::= <QUAL_SCALE>
```

The region area is a set of one or more simple regions, each one defined by one point or a simple polygon (rectangle, square, circle or circle arc).

```
<REGION_AREA> ::= {<SIMPLE_REGION>}

<SIMPLE_REGION> ::= <POINT> | (rect <POINT> <POINT>) | (square <POINT> <POINT>
      <POINT> <POINT>) | (circle <POINT> [real]) | (arc <POINT> [real] [real] [real]
      [real])

<POINT> ::= (point [integer] [integer])
```

Each of the previous messages (**define** and **delete**) should be answered by the application that receive the message:

```
<DEFINE_RESULT> ::= define_result (<ACTION_ID> <ACTION_RESULT>)           [2.3]

<DELETE_RESULT> ::= delete_result (<ACTION_ID> <ACTION_RESULT>)           [2.4]
```

At any time it is possible to know which are the defined regions and their characteristics:

```
<GET_REGIONS> ::= get_region_names                                        [2.5]

<GET_REGION> ::= get_region <REG_NAME>                                     [2.6]
```

The previous messages should obtain as answers, respectively

```
<GET_REGIONS_RESULT> ::= region_names (<ACTION_ID> {<REG_NAME>})          [2.7]

<GET_REGION_RESULT> ::= region (<ACTION_ID> <REG_NAME> <REGION>)          [2.8]
```

There are also messages to get information about the model loaded in the simulator – they can obtain the size, type, morphology and species of molluscs and shellfish prevailing in it:

```
<MODEL_DEFINITIONS> ::= <GET_DIMENSIONS> | <GET_MORPHOLOGY> | <GET_SPECIES> |
        <DIMS_RESULT> | <MORPHOLOGY_RESULTS> | SPECIES_RESULTS>

<GET_DIMENSIONS> ::= model_dimensions                                    [2.9]

<GET_MORPHOLOGY> ::= model_morphology                                    [2.10]

<GET_SPECIES> ::= model_species                                         [2.11]
```

The previous messages should obtain as answers, respectively:

```
<DIMS_RESULT> ::= dimensions (<ACTION_ID> <LINES> <COLUMNS> <LAYERS> <MOD_TYPE>)
                                                                        [2.12]
<MORPHOLOGY_RESULTS> ::= <MORPHOLOGY_RESULT> | <MORPHOLOGY_END>

<MORPHOLOGY_RESULT> ::= morphology (<ACTION_ID> > {(<CELL> [real])})     [2.13]

<MORPHOLOGY_END> ::= morphology_end                                      [2.14]

<SPECIES_RESULTS> ::= <SPECIES_RESULT> | <SPECIES_END>

<SPECIES_RESULT> ::= benthic_species (<ACTION_ID> {(<SPECIES_NAME> <BOXES>)})
                                                                        [2.15]

<SPECIES_END> ::= benthic_species_end                                   [2.16]

<LINES> ::= [integer]

<COLUMNS> ::= [integer]

<LAYERS> ::= [integer]

<MOD_TYPE> ::= 0D | 1DH | 1DV | 2DH | 2DV | 3D

<CELL> ::= [integer]

<SPECIES_NAME> ::= ([string])

<BOXES> → defined in message [3.13] (<GET_VAR_VALUE>)
```

## 2.3 Actions

The actions' messages are closely linked to each type of agent involved in the system.

An agent / application that has an interest in the production of shellfish has actions of deposit, inspect and collect species of molluscs.

```
<ACTION_MSG> ::= <SEED_ACTION> | <INSPECT_ACTION> | <HARVEST_ACTION> | <ACTION_SIM>
```

To deposit (seed), the agent indicates the region, the time, the characteristics of the species of molluscs to deposit and the total weight seeded. The two real values indicated in the message may have different meanings, depending on molluscs in question. By example, for the oysters and scallops, the first value indicates the individual weight of the shell and the second indicates the individual weight of meat; for clams, the first value indicates the individual dry weight, and the second indicates the individual weight.

To inspect, the agent indicates the region and the time of the inspection.

To collect (harvest), it must indicate, beyond the region, the characteristics of shellfish to collect and time of collection.

```
<SEED_ACTION> ::= seed (<REG_NAME> <TIME> <BIVALVE_S> <DENSITY>)          [3.1]

<INSPECT_ACTION> ::= inspect (<REG_NAME> <TIME>)                          [3.2]

<HARVEST_ACTION> ::= harvest (<REG_NAME> <TIME> <BIVALVE>)                [3.3]

<BIVALVE_S> ::= <BTYPE> ([real] [real])

<BIVALVE> ::= <BTYPE> <SHELL_LENGTH>

<BTYPE> ::= scallop | kelp | oyster | mussel | clam

<SHELL_LENGTH> ::= (length [real])

<DENSITY> ::= (density [real])
```

Reference to the action moment can be as quickly as possible (*now*) or one value that indicates the number of seconds from 1970 January, 1 00:00.

```
<TIME> ::= now | [integer]
```

Any agent / application can act over the simulator choosing the model it wants to simulate, controlling the parameterization of the model - gathering / changing parameters of the simulated classes and collecting / recording the results of the simulation. Messages can be divided into four different types:

```
<ACTION_SIM> ::= <MODEL_ACTION> | <EXEC_ACTION> | <SPECS_ACTION> | <REG_ACTION>
```

- Actions to choose the model to simulate - open or close model, survey the model in simulation

```
<MODEL_ACTION> ::= <OPEN_MODEL> | <CLOSE_MODEL> | <GET_MODEL> | <SAVE_CONF>

<OPEN_MODEL> ::= open_model <MODEL_NAME>                                   [3.4]

<CLOSE_MODEL> ::= close_model                                             [3.5]

<GET_MODEL> ::= model_name                                                [3.6]

<SAVE_CONF> ::= save_configuration                                        [3.7]

<MODEL_NAME> ::= [string]
```

- Actions over the simulation execution - monitor and influence the simulations - initialising, running, stopping, restarting and completing the simulations:

```
<EXEC_ACTION> ::= initialise | run | stop | pause | <STEP_CMD>            [3.8]

<STEP_CMD> ::= step [integer]                                             [3.9]
```

- Actions over the model in simulation – to choose / survey classes, choose / inquire initial values for variables and parameters, choose / survey frequency and range of simulation, choose / survey simulation sub-domain:

```
<SPECS_ACTION> ::= <SP_CLASSES> | <SP_VARS> | <SP_PARMS> | <SP_TIME> | <SUB_DOMAIN>

<SP_CLASSES> ::= <GET_CLASSES> | <SELECT_CLASSES>
```

```
<GET_CLASSES> ::= get_available_classes | get_selected_classes          [3.10]

<SELECT_CLASSES> ::= select_classes {<CLASS_NAME>}                       [3.11]

<SP_VARS> ::= <GET_CLASS_VARS> | <GET_VAR_VALUE> | <SET_VAR_VALUE>

<GET_CLASS_VARS> ::= get_variables <CLASS_NAME>                          [3.12]

<GET_VAR_VALUE> ::= get_variable_value <CLASS_NAME> <VAR_NAME> <BOXES>   [3.13]

<SET_VAR_VALUE> ::= set_variable_value <CLASS_NAME> {(<VAR_NAME> <BOXES> [real])}
                                                                        [3.14]

<SP_PARMS> ::= <GET_PARMS> | <SET_PARMS>

<GET_PARMS> ::= get_parameters <CLASS_NAME>                             [3.15]

<SET_PARMS> ::= set_parameters <CLASS_NAME> {(<PARM_NAME> [real])}      [3.16]

<SP_TIME> ::= <GET_TIME> | <SET_TIME>

<GET_TIME> ::= get_time_spec                                           [3.17]

<SET_TIME> ::= set_time_spec <STEP> <START_TIME> <FINISH_TIME>          [3.18]

<SUB_DOMAIN> ::= subdomain <DOMAIN>                                    [3.19]

<CLASS_NAME> ::= ([string])

<VAR_NAME> ::= ([string])

<BOXES> ::= (<SUB_DOMAIN>) | ({<CELL>})

<PARM_NAME> ::= ([string])

<STEP> ::= [integer]

<START_TIME> ::= [integer]

<FINISH_TIME> ::= [integer]

<DOMAIN> ::= all | ({<REG_NAME>})
```

- Actions to record the results - choose variables to register, frequency, range and type of recording, choose sub-domain to register or activate the monitoring mode (trace):

```
<REG_ACTION> ::= <REG_FILE> | <REG_VARS> | <REG_LOG> | <REG_TIME> | <REG_TRACE>

<REG_FILE> ::= output_file <FILE_NAME>                                 [3.20]

<REG_VARS> ::= <GET_VARS> | <SELECT_VARS> | <UNSELECT_VARS>

<GET_VARS> ::= get_available_variables                                 [3.21]

<SELECT_VARS> ::= select_variables <OUTPUT_TYPE> ({<VAR_NAME>}) <BOXES>  [3.22]

<UNSELECT_VARS> ::= unselect_variables <OUTPUT_TYPE> {<VAR_NAME>}        [3.23]

<REG_LOG> ::= log <LOG_TYPE> ({<LOG_STEP>})                             [3.24]

<REG_TIME> ::= <GET_REG_TIME> | <SET_REG_TIME>

<GET_REG_TIME> ::= get_output_time                                     [3.25]

<SET_REG_TIME> ::= set_output_time <STEP> <START_TIME> <FINISH_TIME>    [3.26]

<REG_TRACE> ::= trace                                                  [3.27]
```

```
<FILE_NAME> ::= ([string])

<OUTPUT_TYPE> ::= file | graph | table | remote

<LOG_TYPE> ::= xml | xls | txt | remote

<LOG_STEP> ::= [integer]
```

## *2.4 Perceptions*

The perceptions' messages are closely related with the type of agent involved in the system and also to the actions performed by each over the simulator.

An agent with an interest in the production of shellfish, seed molluscs and its perceptions are the result of the actions taken.

The response to the seed action of the agent may be positive or negative (in the case such action is denied). In response to the inspection action the agent receives a message with the bivalve's characteristics in the region. The resulting harvest is negative or positive, and in this case, it is indicated the total weight harvested.

```
<PERCEPTION_MSG> ::= <SEED_RESULT> | <INSPECT_RESULT> | <HARVEST_RESULT> |
        <PERCEPTION_SIM>

<SEED_RESULT> ::= seed_result (<ACTION_ID> <ACTION_RESULT>)              [4.1]

<INSPECT_RESULT> ::= inspect_result (<ACTION_ID> {<BIVALVE>})           [4.2]

<HARVEST_RESULT> ::= harvest_result (<ACTION_ID> <ACTION_RESULT> <WEIGHT>)   [4.3]

<WEIGHT> ::= [real]
```

The `<ACTION_ID>` of the perception message identifies the `<ID>` of the corresponding action message.

The agents' / applications' perceptions are both messages with the result of the actions initiated by the agent / application or messages spontaneously sent by the simulator. They may be from 5 types:

```
<PERCEPTION_SIM> ::= <MODEL_RESULT> | <EXEC_RESULT> | <SPECS_RESULT> | <REG_RESULT>
        | <EVENT_MSG>
```

- Answers to the actions over the model:

```
<MODEL_RESULT> ::= <OPEN_RESULT> | <CLOSE_RESULT> | <GET_RESULT> | <SAVE_RESULT>

<OPEN_RESULT> ::= open_result (<ACTION_ID> <ACTION_RESULT>)             [4.4]

<CLOSE_RESULT> ::= close_result (<ACTION_ID> <ACTION_RESULT>)           [4.5]

<GET_RESULT> ::= model (<ACTION_ID> <MODEL_NAME>)                       [4.6]

<SAVE_RESULT> ::= save_result (<ACTION_ID> <ACTION_RESULT>)             [4.7]
```

- Answers to the actions over the simulation execution:

```
<EXEC_RESULT> ::= exec_result (<ACTION_ID> <ACTION_RESULT>)            [4.8]
```

- Answers to the actions over the model configuration:

```
<SPECS_RESULT> ::= <CLASSES_RESULT> | <VARS_RESULT> | <PARMS_RESULT> |
      <TIME_RESULT> | <SUB_DOMAIN_RESULT>

<CLASSES_RESULT> ::= <CLASSES_AVAILABLE> | <CLASSES_SELECTED>
<CLASSES_AVAILABLE> ::= classes_available (<ACTION_ID> {<CLASS_NAME>})       [4.9]
<CLASSES_SELECTED> ::= classes_selected (<ACTION_ID> {<CLASS_NAME>})        [4.10]

<VARS_RESULT> ::= <CLASS_VARS> | <VAR_VALUE> | <VAR_SET>
<CLASS_VARS> ::= variables (<ACTION_ID> {<VAR_NAME>})                       [4.11]
<VAR_VALUE> ::= variable_value (<ACTION_ID> {(<CELL> [real])})             [4.12]
<VAR_SET> ::= variable_set_result (<ACTION_ID> <ACTION_RESULT>)           [4.13]

<PARMS_RESULT> ::= <CLASS_PARMS> | <PARMS_SET>
<CLASS_PARMS> ::= parameters_values (<ACTION_ID> {(<PARM_NAME> [real])})   [4.14]
<PARMS_SET> ::= parameters_set_result (<ACTION_ID> <ACTION_RESULT>)       [4.15]

<TIME_RESULT> ::= time_spec (<ACTION_ID> <STEP> <START_TIME> <FINISH_TIME>) [4.16]

<SUB_DOMAIN_RESULT> ::= subdomain_result (<ACTION_ID> <ACTION_RESULT>)     [4.17]
```

- Answers to the actions over the register of the results:

```
<REG_RESULT> ::= <FILE_RESULT> | <REG_VARS_RESULT> | <LOG_RESULT> |
      <REG_TIME_RESULT> | <TRACE_RESULT>

<FILE_RESULT> ::= output_file_result (<ACTION_ID> <ACTION_RESULT>)         [4.18]

<REG_VARS_RESULT> ::= <GET_VARS_RESULT> | <SELECT_VARS_RESULT> |
      <UNSELECT_VARS_RESULT>
<GET_VARS_RESULT> ::= variables_available (<ACTION_ID> {<VAR_NAME>})       [4.19]
<SELECT_VARS_RESULT> ::= select_variables_result (<ACTION_ID> <ACTION_RESULT>)
                                                                          [4.20]
<UNSELECT_VARS_RESULT> ::= unselect_variables_result (<ACTION_ID> <ACTION_RESULT>)
                                                                          [4.21]
<LOG_RESULT> ::= log_result (<ACTION_ID> <ACTION_RESULT>)                 [4.22]
<REG_TIME_RESULT> ::= output_time (<ACTION_ID> <STEP> <START_TIME> <FINISH_TIME>)
                                                                          [4.23]
<TRACE_RESULT> ::= trace_result (<ACTION_ID> <TRACE_STATUS>)              [4.24]
<TRACE_STATUS> ::= on | off
```

- Spontaneous messages sent by the simulator:

```
<EVENT_MSG> ::= <REG_MSG> | <LOG_MSG> | <END_MSG>

<REG_MSG> ::= register (<REG_INDEX> <REG_TIME> <VAR_NAME> {(<CELL> [real])}) [4.25]
<LOG_MSG> ::= logger (<STEP> {(<CLASS_NAME> <FUNC_TYPE> <DATA_CLASS> <VAR_NAME>
      <CELL> [real])})                                                    [4.26]
```

```
<END_MSG> ::= end_simulation | end_step | running | stopped | paused        [4.27]

<REG_INDEX> ::= [integer]

<REG_TIME> ::= [integer]

<FUNC_TYPE> ::= Inquiry | Update

<DATA_CLASS> ::= <CLASS_NAME>
```

# 3 Communications Protocol

The communication between the simulator (EcoDynamo application) and the other actors present in the simulation system is usually of the type handshake - a message-type action expects to receive an answer from the destination application; that response comes in the form of a perception type message.

Only the spontaneous messages and the logging results sent by the simulator don't require feedback.

The first message of each agent for the simulator must be connected (connect). The reception of a positive acceptance message (to accept ok result) indicates that the agent was registered in the simulator as an agent interested in obtaining results from the simulations. When the agent leaves the system it must send the message to disconnect from the simulator.

The simulator, before leaving the system, must send to all registered active agents the disconnect message.

The communications' protocol establishes the answers expected for each action, according to the following table:

| [msg] | Message | Expected answer | [msg] |
|-------|---------|-----------------|-------|
| [1.1] | connect | accept | [1.3] |
| [1.2] | disconnect | accept | [1.3] |
| [1.4] | agents | known_agents | [1.5] |
| [2.1] | define | define_result | [2.3] |
| [2.2] | delete | delete_result | [2.4] |
| [2.5] | get_region_names | region_names | [2.7] |
| [2.6] | get_region | region | [2.8] |
| [2.9] | model_dimensions | dimensions | [2.12] |
| [2.10] | model_morphology | morphology [1] | [2.13] |
| [2.10] | model_morphology | morphology_end [2] | [2.14] |
| [2.11] | model_species | benthic_species [3] | [2.15] |
| [2.11] | model_species | benthic_species_end [4] | [2.16] |
| [3.1] | seed | seed_result | [4.1] |

---

[1] This is the answer while there were messages to send from morphology: morphology of each message has, at most, 750 elements.
[2] This is the answer indicating end of morphology messages.
[3] This is the answer while there were messages to send from benthic species: each benthic species message has, at most, 150 elements.
[4] This is the answer indicating end of benthic species messages.

| [msg] | Message | Expected answer | [msg] |
|-------|---------|-----------------|-------|
| [3.2] | inspect | inspect_result | [4.2] |
| [3.3] | harvest | harvest_result | [4.3] |
| [3.4] | open_model | open_result | [4.4] |
| [3.5] | close_model | close_result | [4.5] |
| [3.6] | model_name | model | [4.6] |
| [3.7] | save_configuration | save_result | [4.7] |
| [3.8] | initialise | exec_result | [4.8] |
| [3.8] | run | exec_result | [4.8] |
| [3.8] | stop | exec_result | [4.8] |
| [3.8] | pause | exec_result | [4.8] |
| [3.9] | step | exec_result | [4.8] |
| [3.10] | get_available_classes | classes_available | [4.9] |
| [3.10] | get_selected_classes | classes_selected | [4.10] |
| [3.11] | select_classes | classes_selected | [4.10] |
| [3.12] | get_variables | variables | [4.11] |
| [3.13] | get_variable_value | variable_value | [4.12] |
| [3.14] | set_variable_value | variable_set_result | [4.13] |
| [3.15] | get_parameters | parameters_values | [4.14] |
| [3.16] | set_parameters | parameters_set_result | [4.15] |
| [3.17] | get_time_spec | time_spec | [4.16] |
| [3.18] | set_time_spec | time_spec | [4.16] |
| [3.19] | subdomain | subdomain_result | [4.17] |
| [3.20] | output_file | output_file_result | [4.18] |
| [3.21] | get_available_variables | variables_available | [4.19] |
| [3.22] | select_variables | select_variables_result | [4.20] |
| [3.23] | unselect_variables | unselect_variables_result | [4.21] |
| [3.24] | log | log_result | [4.22] |
| [3.25] | get_output_time | output_time | [4.23] |
| [3.26] | set_output_time | output_time | [4.23] |
| [3.27] | trace | trace_result | [4.24] |
| [4.25] | register | -- | |
| [4.26] | logger | -- | |
| [4.27] | end_simulation | -- | |
| [4.27] | end_step | -- | |
| [4.27] | running | -- | |
| [4.27] | stopped | -- | |

| [msg] | Message | Expected answer | [msg] |
|-------|---------|-----------------|-------|
| `[4.27]` | paused | -- | |

# 4 Software

The code with the implementation of the communications protocol is provided in one C++ DLL (Dynamic Link Library) that can be imported by any application. The DLL is part of the package distribution of the **EcoDynamo** (simulation application).

## 4.1 Header Files

The header files contain the definition of the `EcoDynProtocol` class, the message symbols and the data structures used.

<u>Folder</u>: `DLLs/ECDProtocol`
<u>Files</u>:   `EcoDynProtocol.h,   ECDPMessages.h,   ECDPAgents.h,   AgentsTable.h   e Region.h.`
<u>Note</u>: the file `EcoDynProtocol.h` includes the other ones.

## 4.2 Library File

<u>Folder</u>: `DLLs/ECDProtocol/Lib`
<u>File</u>: `ECDP.lib`

## 4.3 Dll File

<u>Folder</u>: `DLLs/ECDProtocol/Lib`
<u>File</u>: `ECDP.dll`

## 4.4 Dependencies

The library file with the protocol code uses the `Queue` and `Parser` classes that belongs to the `Utilities` library.

<u>Folder</u>: `DLLs/Utilities/Lib`
<u>File</u>: `Utilities.dll`

<u>Folder</u>: `DLLs/Utilities`
<u>Header Files</u>: `Queue.h, Parser.h`

# 5 Document Versions

## Version 1.3c

Date: 14.May.2008

Messages added:

   - **agents**: ask the known agents;

   - **known_agents**: list with the known agents.

### Version 1.3b

Date: 03.July.2007

Messages added:

- **save_configuration**: saves the actual configuration of the model in the simulator;

- **save_result**: confirms the save operation in the simulator.

### Version 1.3a

Date: 15.May.2007

Messages added:

- **morphology_end**: end of morphology messages;

- **benthic_species_end**: end of benthic species message.

### Version 1.3

Date: 30.April.2007

Messages **get_variable_value** and **variable_value** changed.

Messages added:

- **get_region_names** and **region_names**: regions defined in the application;

- **get_region** and **region**: region characteristics;

- **model_dimensions** and **dimensions**: dimensions of the simulation model;

- **model_morphology** and **morphology**: morphology of the simulation model

- **model_species** and **benthic_species**: benthic species of the simulation model.

### Version 1.2

Date: 15.March.2007

Messages added: **register** and **logger**.

### Version 1.1

Date: 26.February.2007

Messages added:

- **running**: simulation running;

- **stopped**: simulation aborted / stopped;

- **paused**: pause in simulation.

### Version 1.0

Date: 29.January.2007

Messages added:

- **delete**: remove one defined region;

- **define_result** and **delete_result**: result of define and delete actions;

- **unselect_variables**: unselect variables from register values;

- **unselect_variables_result**: result of the previous action.

### Version 0.4a

Date: 27.June.2006

Messages changed:

- **seed**: type `<BIVALVE>` replaced by a new type `<BIVALVE_S>` that includes two real values (bivalve type can change their meaning);

- **step**: one more field included to indicate how many steps to run.

Included two new spontaneous messages sent by the simulator indicating end of step or end of simulation (**end_step** and **end_simulation**).

### Version 0.4

Date: 16.May.2006

Included two new messages to enable the connection from several agents and applications in the same system or computer.

New group included:

- group `<CONNECTION_MSG>` with messages: **connect**, **disconnect, accept**.

### Version 0.3g

Date: 23.January.2006

Type **clam** added to the variable `<BTYPE>`.

### Version 0.3f

Date: 21.December.2005

Changed the message **select_variables**: type **remote** included in `<OUTPUT_TYPE>` and `<SUB_DOMAIN>` replaced by `<BOXES>`.

Changed the message **log**: type **remote** included in `<LOG_TYPE>`.

### Version 0.3e

Date: 21.October.2005

Changed the definition of the messages **seed** and **harvest**.

### Version 0.3d

Date: 03.October.2005

Changed the definition of **point**. Fields are now integers.

### Version 0.3c

Date: 27.June.2005

Changed the messages:

- **variable_value**: removed the fields `<VAR_NAME>` and `<CELL>`;

- **trace_result**: included field `<ACTION_ID>`;

- corrected field `<REG_TIME>`.

Chapters renumbered and protocol table included.

### Version 0.3b

Date: 05.May.2005

Changed the message **parameters_value** to **parameters_values**. Included **mussel** in the `<BTYPE>` definition.

- **variable_value**: removed the fields `<VAR_NAME>` and `<CELL>`;

- **trace_result**: included field `<ACTION_ID>`;

- corrected field `<REG_TIME>`.

Chapters renumbered and protocol table included.

### Version 0.3

Date: 11.February.2005

Included the groups `<ACTION_SIM>` and `<SIMULATION_SIM>`.

### Version 0.2

Date: 27.May.2004

Document reformulation.

### Version 0.1

Date: 21.May.2004

Initial version.

# 6 References

Backus, J.W., Bauer, F.L., Green, J., Katz, C., McCarthy, J., Perlis, A.J., Rutishauser, H., Samelson, K., Vauquois, B., Wegstein, J.H., Wijngaarden, A.v. and Woodger, M. (1960). Report on the algorithmic language ALGOL 60. *Commun. ACM*. 3:5. p. 299-314.

Pereira, A. and Duarte, P. (2005). *EcoDynamo - Ecological Dynamics Model Application*. Porto: University Fernando Pessoa.

Pereira, A., Duarte, P. and Reis, L.P. (2005). *ECOLANG - A communication language for simulations of complex ecological systems*. Riga, Latvia.

Reis, L.P. and Lau, N. (2002). *COACH UNILANG - a Standard Language for Coaching a (Robo)Soccer Team*. In: Robocup 2001 - Robot Soccer World Cup, Seattle, WA, USA.